\documentclass[runningheads]{llncs}
\usepackage[T1]{fontenc}
\usepackage{graphicx}
\usepackage{booktabs}
\usepackage[misc]{ifsym}
\newcommand{\corr}{(\Letter)}

\usepackage{amsmath}
\usepackage{amssymb}
\usepackage{hyperref}
\usepackage{url}
\usepackage{subcaption}
\usepackage{pifont} 

\newcommand{\x}{\ensuremath{\mathbf{x}}} 
\newcommand{\xt}[1][t]{\ensuremath{\x_{#1}}} 
\newcommand{\h}{\ensuremath{\mathbf{h}}} 
\newcommand{\htt}[1][t]{\ensuremath{\h_{#1}}} 
\newcommand{\y}{\ensuremath{\mathbf{y}}} 
\newcommand{\yt}[1][t]{\ensuremath{\y_{#1}}} 

\newcommand{\A}{\ensuremath{\mathbf{A}}} %
\newcommand{\Loss}{\ensuremath{\mathcal{L}}} 

\newcommand{\params}{\ensuremath{\boldsymbol{\theta}}} 
\newcommand{\cell}{\ensuremath{f_{\params}}} 

\newcommand{\pparams}{\ensuremath{\boldsymbol{\phi}}} 
\newcommand{\pred}{\ensuremath{g_{\pparams}}} 

\newcommand{\cmark}{\ding{51}}%
\newcommand{\xmark}{\ding{55}}%

\begin{document}

\title{Fast Training of Recurrent Neural Networks with Stationary State Feedbacks}

\titlerunning{Fast Training of RNNs with Stationary State Feedbacks}

\author{Paul Caillon\inst{1}\textsuperscript{†} \corr \and Erwan Fagnou\inst{1}\textsuperscript{†} \and Alexandre Allauzen\inst{1,2}}


\authorrunning{P. Caillon \and E. Fagnou \and A. Allauzen}
 \institute{Miles Team, LAMSADE, Université Paris Dauphine - PSL, Paris, France \and ESPCI PSL, Paris, France \\ \email{\{name.surname\}@dauphine.psl.eu}}

\maketitle              
\renewcommand{\thefootnote}{}\footnotetext{\textsuperscript{†}Equal contribution.}
\setcounter{footnote}{0} 
\renewcommand{\thefootnote}{\arabic{footnote}}  
\begin{abstract}
Recurrent neural networks (RNNs) have recently demonstrated strong performance and faster inference than Transformers at comparable parameter budgets. However, the recursive gradient computation with the backpropagation through time (or BPTT) algorithm remains the major computational bottleneck. In this work, we propose a novel method that replaces BPTT with a fixed gradient feedback mechanism, yielding an efficient approximation of the exact gradient propagation based on the assumption of time stationarity.
Our approach leverages state-space model (SSM) principles to
define a structured feedback matrix that directly propagates
gradients from future time steps. This formulation bypasses the
need for recursive gradient backpropagation, significantly
reducing training overhead while preserving the network's
ability to capture long-term dependencies. The experiments on
language modeling benchmarks exhibit competitive perplexity
scores, while significantly reducing the training costs. These
promising results suggest that designing a feedback method like an SSM can fully exploit the efficiency advantages of RNNs for many practical applications.

\end{abstract}

\section{Introduction}

For many decades, recurrent neural networks (RNNs) have played a key role in
sequence modeling~\cite{hopfield1982neural,jordan1997serial}, from
natural language processing to DNA sequence classification and from
task planning to decision-making, to give a few examples.  By design,
recurrent architectures explicitly embed the time in their
structure~\cite{elman1990finding}.  Similar to a discrete-time
dynamical system, an RNN unrolls a structure named the recurrent cell to
memorize a sequence of arbitrary length in an online fashion.
Focusing on the design of the cell, sophisticated
architectures like LSTM~\cite{hochreiter1998vanishing} and GRU~\footnote{\textit{Long Short Term
    Memory} and \textit{Gated Recurrent Unit}}~\cite{cho2014learning} yielded important
improvements in the early 2000 for tasks like handwritten text recognition \cite{graves2008offline} and later 
 machine translation with the introduction of attention~\cite{bahdanau2014neural}. 
 However their prohibitive training time has limited their scalability
 and widespread use.  Indeed, the backpropagation through time
 (BPTT) algorithm used to compute the gradients cannot be parallelized,
 precisely due to the recurrent structure.

 This issue was an important motivation for the development of
 transformers~\cite{vaswani2017attention}. This new architecture was a
 huge breakthrough and deeply changed the research
 landscape. With their unprecedented scalability, transformers allow
 the training of sequence models (\textit{e.g.} the foundation models)
 of sizes that were out of reach with recurrent models.  Nevertheless
 they trade the training speed for new limitations like the quadratic
 complexity in the input length, or their inefficiency to process
 inputs in an online scenario. Therefore the pendulum recently swung
 back a bit with a regain of interest for recurrent architectures
 \cite{orvieto2023resurrecting,beck2024xlstm} and state space models
 \cite{Gu22S4,gu2024mambalineartimesequencemodeling}.  While these new
 models can achieve competitive results at comparable parameter
 budgets, this recent line of research focuses on simplifying the
 recurrence to speed up inference and training, with the drawback of
 harming the memorization capacity.

 In this paper, we address the prohibitive training time from a
 different standpoint.  Since the bottleneck is the BPTT, we introduce
 a training algorithm for recurrent architecture that efficiently
 approximates the gradient computation. More precisely, the gradients
 can be decomposed into two terms to be computed. The first one is
 local and time independent, while the second one corresponds to the
 backward propagation in time. The bottleneck arises in the second
 term, but this is specifically the heart of the memorization
 process. Therefore we propose in this work to simplify this term,
 while keeping the error backpropagation through time.

 Our main contributions are the following:
 \begin{itemize}
 \item We extend the idea of Direct Feedback Alignment
   \cite{nokland2016direct} to recurrent architecture: the
   time-dependent Jacobian matrix of the recurrent cell involved in
   gradient can be approximated by a random and fixed matrix. This
   allows us to efficiently propagate gradients from all the output
   layers to hidden layers.
 \item By further assuming that this approximation is stationary in
   time, the BPTT can be rewritten as a Linear Time Invariant
   dynamical system that conveys the error propagation backward in time.
 \item In a nutshell, our algorithm replaces the BPTT by the inference
   of a state space model (SSM) in reversed time. We can then benefit
   from the fast inference of SSMs as well as their ability to handle
   long range dependencies.
 \end{itemize}

\section{Gradient Computation for Recurrent Networks}\label{sec:bptt}


Let us consider a recurrent neural network defined by the following recurrence relation:
\begin{equation}
    \htt = \cell(\htt[t-1], \xt), \quad \yt = \pred(\htt). 
\end{equation}
In this equation, $\htt$ denotes the hidden state vector at time step
$t$, which is estimated by the recurrent cell $\cell(\cdot,\cdot)$
parameterized by $\params$. We will alternatively write $\htt = \cell^t(\htt[t-1])$ to simplify notations.
Given the input sequence of observations
$(\xt)_{t=1}^{T}$, the recurrent network acts like a transducer and
generates at each time step the output $\yt$ using $\pred$. This
prediction function maps hidden states to outputs.  In a supervised
setting, this model can be trained in order to minimize the loss
function $\Loss_t = l(\yt, \yt^*)$ at every time step $t$ (typically
the negative log-likelihood).  The total loss over the sequence is
defined by:
\[
\Loss = \sum_{t=1}^{T} \Loss_t.
\]

\subsection{Backpropagation through time}
\label{ssec:bptt}
Training recurrent neural networks with gradient descent involves computing gradients of the total loss $\Loss$ with respect to the parameters $\params$ via BPTT:
\begin{align}
    \frac{\partial \Loss}{\partial \params}
    &= \sum_{t=1}^{T} \frac{\partial \Loss_t}{\partial \params}, \quad\text{with}\quad
    \frac{\partial \Loss_t}{\partial \params} 
    = \underbrace{\frac{\partial \Loss_t}{\partial \yt}\frac{\partial \yt}{\partial \htt}}_{\mathbf{e}_t} \frac{\partial \htt}{\partial \params}.
\end{align}

The error term $\mathbf{e}_t$ corresponds to the feedback received
from the prediction layer at time $t$. Computing
$\frac{\partial \Loss_t}{\partial \params}$ explicitly involves
recursively unfolding the recurrent state through time because each
hidden state $\htt$ depends on previous hidden states:

\begin{align}
    \frac{\partial \Loss_t}{\partial \params} 
    &= \mathbf{e}_t \frac{\partial \htt}{\partial \params} \\
    &= \mathbf{e}_t \left(
        \frac{\partial \cell^t}{\partial \params} 
        + \frac{\partial \htt}{\partial \htt[t-1]} \frac{\partial \htt[t-1]}{\partial \params}
    \right) \\
    &= \mathbf{e}_t\left(\frac{\partial \cell^t}{\partial \params} 
        + \frac{\partial \htt}{\partial \htt[t-1]}\left(\frac{\partial \cell^{t-1}}{\partial \params}
        + \frac{\partial \htt[t-1]}{\partial \htt[t-2]}\frac{\partial \htt[t-2]}{\partial \params}\right)
    \right),\quad \text{and so forth.}
\end{align}

Generalizing this expression, we obtain the standard form of BPTT:

\begin{align}\label{eq:bptt}
    \frac{\partial \Loss_t}{\partial \params} 
    &= \mathbf{e}_t \sum_{k=1}^{t}\left(
    \prod_{j=t}^{k+1}\frac{\partial \htt[j]}{\partial \htt[j-1]}\right)\frac{\partial \cell^k}{\partial \params}.
\end{align}

The product inside the summation represents the backward propagation of the gradient through the hidden states, from the time step $t$ to the time step $k$.

\subsection{Recurrent gradient equation}
However, the gradients computed by BPTT can also be expressed through an alternative formulation, explicitly highlighting the recursive structure of the gradient computation with respect to the hidden states $\htt$. Considering the total derivative of the loss, we obtain:

\begin{align}
  \frac{\partial \Loss}{\partial \params} 
  &= \sum_{t=1}^{T} \frac{\partial \Loss}{\partial \htt}\frac{\partial \cell^t}{\partial \params},\\[0.5em]
  \frac{\partial \Loss}{\partial \htt} 
  &= \sum_{k=t}^{T} \frac{\partial \Loss_k}{\partial \htt}
   = \underbrace{\frac{\partial \Loss_t}{\partial \htt}}_{\mathbf{e}_t}
   + \sum_{k=t+1}^{T}\underbrace{\frac{\partial \Loss_k}{\partial \htt[t+1]}\frac{\partial \htt[t+1]}{\partial \htt}}_{\text{chain rule}} \\[0.5em]
  &= \mathbf{e}_t + \left(\sum_{k=t+1}^{T}\frac{\partial \Loss_k}{\partial \htt[t+1]}\right)\frac{\partial \htt[t+1]}{\partial \htt} \\[0.5em]
  &= \mathbf{e}_t + \frac{\partial \Loss}{\partial \htt[t+1]} \frac{\partial \htt[t+1]}{\partial \htt}.
\end{align}

Introducing the following notations for the gradient term and the Jacobian matrix of the recurrent cell:
\[
\mathbf{g}_t := \frac{\partial \Loss}{\partial \htt} \quad \A_t := \frac{\partial \htt[t+1]}{\partial \htt},
\]

we can derive  a concise recurrence relation for the gradient computation that takes the form of a standard Initial Value Problem, but reversed in time:
\begin{equation}\label{eq:bptt-ssm}
  \mathbf{g}_t = \mathbf{e}_t + \mathbf{g}_{t+1} \A_t,\quad\text{with "initial" condition}\quad \mathbf{g}_T = \mathbf{e}_T.
\end{equation}

This last equation describes a linear dynamical system that evolves backwards with time, where the final gradient $\mathbf{g}_T$ plays
the role of an initial state.  It is worth noticing that the Jacobian
matrix $\A_t$ explicitly depends on time through the hidden
states. Nevertheless, aggregating over all time steps yields a compact
expression for the gradient with respect to parameters:

\begin{equation}
\frac{\partial \Loss}{\partial \params} = \sum_{t=1}^{T} \mathbf{g}_t \frac{\partial \cell^t}{\partial \params}.
\end{equation}

In this formulation, the recursive gradient computation is entirely encapsulated within the terms $\mathbf{g}_t$, estimated by traversing the sequence backward in time. In terms of dimension, we have:
\begin{itemize}
    \item $\mathbf{g}_t$ and $\mathbf{e}_t$ as row vectors in $\mathbb{R}^{d}$;
    \item $\A_t$ as a matrix in $\mathbb{R}^{d \times d}$;
    \item $\frac{\partial \cell^t}{\partial \params}$ as a matrix in $\mathbb{R}^{N_{\params} \times d}$.
\end{itemize}
This formulation allows us to propose simplifying assumptions in order to design a more scalable training algorithm. 

\section{Diagonal State Feedbacks}\label{sec:dsf}

In standard BPTT, the gradient
computation involves recursively multiplying Jacobian matrices,
denoted by
$\A_t = \frac{\partial \mathbf{h}_{t+1}}{\partial \mathbf{h}_{t}}$ in
the previous section.  This matrix gathers the partial derivatives of
the recurrent cell with respect to $\htt$.  Although the parameters
$\params$ of the cell are tied across the time steps, the Jacobian
matrix is therefore time-dependent since it corresponds to the derivatives
(the same function) measured at different points $\htt$ for every $t$.
As a consequence, the computational cost increases drastically with
sequence length, both in terms of time complexity and memory
requirements.

In this paper, we propose an alternative formulation of the gradient
computation by approximating the time-dependent Jacobian $\A_t$ with a
single and time-invariant feedback matrix $\A$. This approximation
means that we consider a stationary dynamic for the hidden-state
transition, \textit{i.e.}, constant across all time steps.
Under this assumption, the backward gradient propagation defined by
equation~\eqref{eq:bptt-ssm} simplifies significantly into a linear,
and time-invariant dynamical system operating in reverse time:

\begin{equation}\label{eq:dfa-ssm}
\left\{
\begin{aligned}
    \mathbf{g}_{T} &= \mathbf{e}_{T},\\[0.3em]
    \mathbf{g}_{t} &= \mathbf{e}_{t} + \mathbf{g}_{t+1}\mathbf{A}, \quad t<T.
\end{aligned}
\right.
\end{equation}

The matrix $\mathbf{A}$ acts as a stationary feedback mechanism, consistently propagating gradients from future time steps back to the current state.
As already described in the literature associated with SSMs \cite{Gu22S4}, this recursive equation can be formulated as a convolution:
\begin{equation}
(\mathbf{g}_T, \mathbf{g}_{T-1},\dots,\mathbf{g}_1) 
= \underbrace{(\mathbf{e}_T,\mathbf{e}_{T-1},\dots,\mathbf{e}_1)}_{\text{input errors}} *
\underbrace{(\mathbf{I}, \mathbf{A}, \mathbf{A}^2, \dots, \mathbf{A}^{T-1})}_{\text{convolution kernel}}\,.
\end{equation}
This peculiar convolution can be efficiently computed in the Fourier
domain. For efficiency concerns, we constrain the matrix $\A$'s structure to ease the pre-computation of the convolution kernel as discussed in the following section~\ref{sec:DSFapp}.

\subsection{Diagonal State Feedbacks (DSF) Approximation}
\label{sec:DSFapp}
If $\A$ is an arbitrary matrix, sequentially raising it to the
power $t$ is computationally intensive, with a complexity in
$\mathcal{O}(d^3)$ per multiplication. Recent works on State Space Models
(SSMs) have proposed introducing structured constraints on the
transition matrix to reach better computational efficiency and
stability. Specifically, several structured forms have been
introduced, including:

\begin{itemize}
\item \textbf{HiPPO matrices (High-order Polynomial Projection
    Operators)}~\cite{gu2020hippo}, which leverage polynomial
  projection operators to capture structured long-range memory,
  facilitating efficient memory representation and computation in
  gradient propagation.
\item \textbf{Structured State Space models (S4 and
    S5)}~\cite{Gu22S4,smith2022simplified}, which build upon HiPPO
  operators but propose a Normal Plus Low-Rank.  This approximation
  admits fast implementations, improving numerical stability,
  computational complexity, and memory efficiency.
\item \textbf{Diagonal State Space models} \cite{gu2022s4d,gupta2022diagonal,ma2023mega} further simplify the structure of $\A$, by
  assuming a diagonal form. They show experimentally that, with this
  approximation, SSMs can achieve the same level of performance.
\end{itemize}

Drawing inspiration from these structured approaches, our method
relies on a diagonal feedback matrix $\mathbf{A}$.  We therefore refer
to this method as the \emph{Diagonal State Feedbacks (DSF)}
approximation. This choice significantly reduces computational
complexity down to merely $\mathcal{O}(d)$ per time step. In addition,
storing a diagonal feedback matrix requires memory complexity of
$\mathcal{O}(d)$ rather than $\mathcal{O}(d^2)$, enhancing scalability
for large hidden-state dimensions.
In comparison with the BPTT, our proposed DSF approximation represents an extreme
simplification, achieving computational efficiency without sacrificing empirical performance.

\subsection{Fixed Feedback Matrix: Connection to Direct Feedback Alignment}

Our Diagonal State Feedbacks method closely relates to Direct Feedback
Alignment (DFA), originally introduced
by~\cite{nokland2016direct}. DFA employs a fixed and randomly
initialized feedback matrix to propagate gradients from output layers
to hidden layers, thereby removing the need to use the exact Jacobians. However, in contrast to DFA, our approach applies
this principle specifically to recurrent network gradient propagation
across time rather than across layers.

Precisely, we fix the diagonal feedback matrix $\mathbf{A}$ at
initialization, completely avoiding the gradient-based optimization
that is necessary for structured matrices in standard SSMs such as S4
or HiPPO. By keeping $\mathbf{A}$ constant during training, we remove
the computational overhead associated with optimizing feedback matrix
parameters and simplify the backward gradient computation drastically.
This approach thus presents several advantages:
\begin{itemize}
\item \textbf{Computational efficiency}: Each gradient step only uses
 element-wise operations, reducing the computational complexity from
  $\mathcal{O}(d^2)$ (cost of a vector-Jacobian product) to $\mathcal{O}(d)$.
\item \textbf{Memory efficiency}: Only $\mathcal{O}(d)$ memory is
  required to store the diagonal feedback matrix.
\item \textbf{Stability}: By fixing the diagonal
  feedback structure, we eliminate numerical instabilities arising from the dynamic
  Jacobian computation. It allows us to readily mitigate the exploding and/or vanishing gradient
  issues. 
\item With this improved \textbf{scalability}, large scale training of
  recurrent networks with wide hidden state dimensions becomes
  possible, even for very long sequences.
\end{itemize}

Our assumptions on time-invariance and structural simplicity may appear as drastic. However, the 
empirical results demonstrate that the proposed Diagonal State
Feedbacks method achieves competitive performance compared to
standard BPTT, validating our hypothesis that the complexity
introduced by fully time-dependent or structured Jacobians may not be
essential for learning effective recurrent representations.

\section{Computational Complexity Analysis}
\label{sec:complexity}

We now analyze the computational complexity and sequentiality of gradient computations for the standard BPTT and our proposed DSF method.
We summarize the complexity results discussed below in Table~\ref{tab:complexity_summary}.
\begin{table}[ht]
\caption{Complexity and Sequentiality Comparison}\label{tab:complexity_summary}
\centering
\begin{tabular}{lcc}
\toprule
\textbf{Method} & \textbf{Complexity} & \textbf{Sequentiality} \\
\midrule
BPTT & $\mathcal{O}(d^2 T)$ & $\mathcal{O}(T)$ \\
DSF (naive implementation) & $\mathcal{O}(d\,T)$ & $\mathcal{O}(T)$ \\
DSF (FFT/Prefix-sum implementation) & $\mathcal{O}(d\,T\log T)$ & $\mathcal{O}(\log T)$ \\
Fully Truncated BPTT (FT-BPTT) & $\mathcal{O}(1)$ & $\mathcal{O}(1)$ \\
\bottomrule
\end{tabular}
\end{table}
This table illustrates how the proposed DSF approach offers substantial computational benefits compared to standard BPTT, especially in structured implementations (FFT or prefix-sums). DSF provides a favorable balance between computational efficiency, memory usage, and sequential complexity, making it attractive for large-scale recurrent network training.

\subsection{Complexity of Backpropagation Through Time (BPTT)}

Recall that training RNNs via standard BPTT involves recursively computing gradients through the network's hidden states (see Section~\ref{sec:bptt}). At each time step $t$, the gradient $\mathbf{g}_t$ w.r.t the hidden state is computed as:
$\mathbf{g}_t = \mathbf{e}_t + \mathbf{g}_{t+1}\mathbf{A}_t, \, \mathbf{g}_T = \mathbf{e}_T$.

Here, $\mathbf{A}_t$ is the Jacobian of the recurrent cell, $\frac{\partial \mathbf{h}_{t+1}}{\partial \mathbf{h}_t}$, and it typically varies with time and input. Evaluating this vector-Jacobian multiplication at each time step has a computational complexity of $\mathcal{O}(d^2)$. Repeating this operation across all time steps leads to a total complexity of $\mathcal{O}(d^2 T)$.
Moreover, due to its inherently recursive definition, this computation has a sequential dependency across all time steps, thus sequentiality scales linearly with the sequence length $T$.

\subsection{Diagonal State Feedbacks (DSF)}

In our proposed Diagonal State Feedbacks (DSF) approximation, we assume the Jacobian matrix $\mathbf{A}_t$ is replaced by a fixed, diagonal matrix $\mathbf{A}$ throughout training. Consequently, the gradient recursion simplifies to a stationary diagonal multiplication:
$\mathbf{g}_t = \mathbf{e}_t + \mathbf{g}_{t+1}\mathbf{A}, \, \text{with } \mathbf{A} \text{ diagonal}$.
Since $\mathbf{A}$ is diagonal, each recursive gradient update involves only element-wise operations. This reduces the complexity per time step from $\mathcal{O}(d^2)$ to $\mathcal{O}(d)$, cumulating to $\mathcal{O}(d\,T)$.
However, this computation remains sequential, as each gradient update at time $t$ depends on the next step $t+1$, leading to sequentiality of $\mathcal{O}(T)$.

\subsection{Reducing Sequentiality with Prefix-Sums or FFT}

Because $\mathbf{A}$ is diagonal and fixed, the recursion:
$\mathbf{g}_t = \mathbf{e}_t + \mathbf{A}\mathbf{e}_{t+1} +  \dots + \mathbf{A}^{T-t}\mathbf{e}_T$,
can be written as a discrete convolution operation in reverse-time order. Leveraging efficient prefix-sum (see \cite{blelloch1990prefixsums}, \cite{cumsum1986hillis}) or Fast Fourier Transform (FFT) algorithms, we reduce sequentiality dramatically to logarithmic complexity $\mathcal{O}(\log T)$. Such approaches rely on parallelization techniques (prefix-scan algorithms or FFT-based convolutions), leading to a complexity of $\mathcal{O}(d\,T\log T) $,
while reducing sequentiality to $\mathcal{O}(\log T)$.
Thus, DSF can be efficiently implemented on parallel hardware like GPUs, achieving superior practical runtime performance.

\subsection{Fully Truncated BPTT (No Temporal Feedback)}

A further simplification is the Fully Truncated BPTT (FT-BPTT), where the gradient computation completely neglects the backward propagation through time, \textit{i.e.}, the state transitions' Jacobian is set to zero. In this extreme approximation, each gradient update depends solely on the current time step's error signal, effectively treating each time step independently. Hence, the complexity and sequentiality reduce drastically to $\mathcal{O}(1)$ each.

This approach, while computationally efficient, disregards temporal dependencies entirely and typically results in degraded learning performance.

\section{Experimental Results and Analysis}
\subsection{Experimental Setup}
We conduct extensive experiments to assess the effectiveness of the proposed Diagonal State Feedbacks (DSF) method in comparison to standard BPTT and FT-BPTT. The experiments cover various configurations of recurrent architectures, dataset sizes, and model complexities. The exact experimental details are given in the supplementary material. The code is available on \href{https://github.com/p0lcAi/DSF}{Github}.

We conduct experiments on two standard language modeling benchmarks: Penn Treebank (PTB) and Wikitext-103. We compare the performance of DSF against BPTT -- the exact gradient computation method -- and FT-BPTT, which truncates all temporal dependencies. Doing better than FT-BPTT is a necessary condition for DSF to be considered effective, as it should at least capture some temporal dependencies.

On WikiText-103, the model follows a transformer-like architecture, with the attention layers replaced by RNNs. Unless stated otherwise, we use 3 GRU layers with 512 hidden units.

The diagonal feedback matrix $\mathbf{A}$ is initialized using a uniform distribution in the range $[0, 1]$. While we found this approach to be the most successful, future work may attempt to find more suitable initialization strategies.

\subsection{Impact of Network Depth and Width on Penn Treebank}

\begin{figure}[ht]
    \centering
    \begin{subfigure}{0.48\textwidth}
        \centering
        \includegraphics[width=\linewidth]{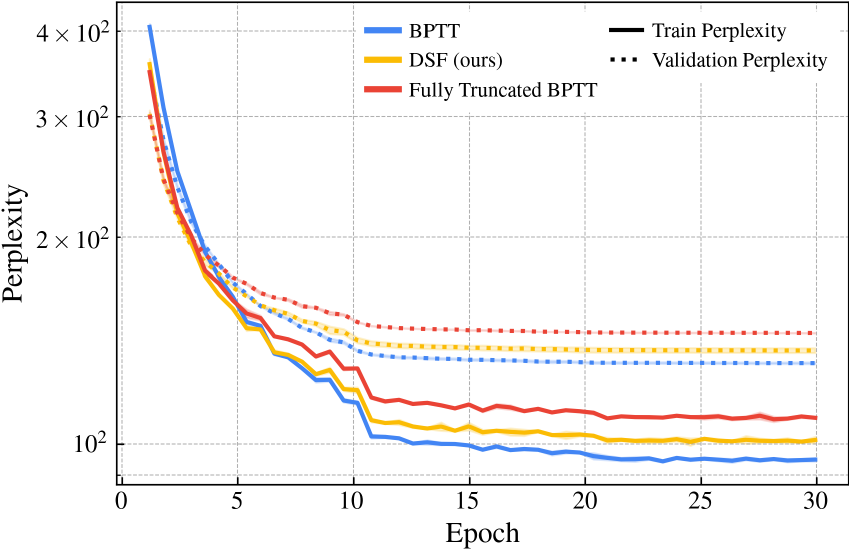}
        \caption{1 GRU Layer with 256 Hidden Units}
        \label{fig:ptb1L256}
    \end{subfigure}
    \hfill
    \begin{subfigure}{0.48\textwidth}
        \centering
        \includegraphics[width=\linewidth]{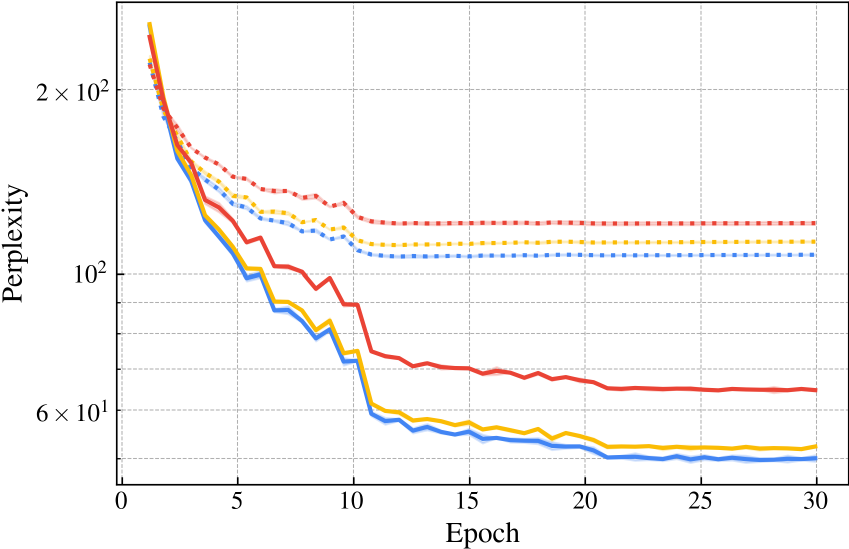}
        \caption{3 GRU Layers with 256 Hidden Units}
        \label{fig:ptb3L256}
    \end{subfigure}

    \begin{subfigure}{0.48\textwidth}
        \centering
        \includegraphics[width=\linewidth]{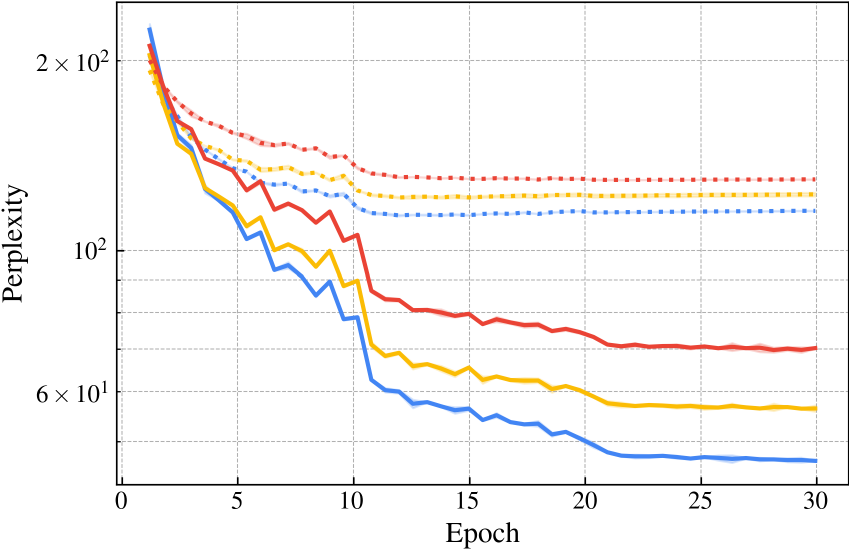}
        \caption{1 GRU Layer with 1024 Hidden Units}
        \label{fig:ptb1L1024}
    \end{subfigure}
    \hfill
    \begin{subfigure}{0.48\textwidth}
        \centering
        \includegraphics[width=\linewidth]{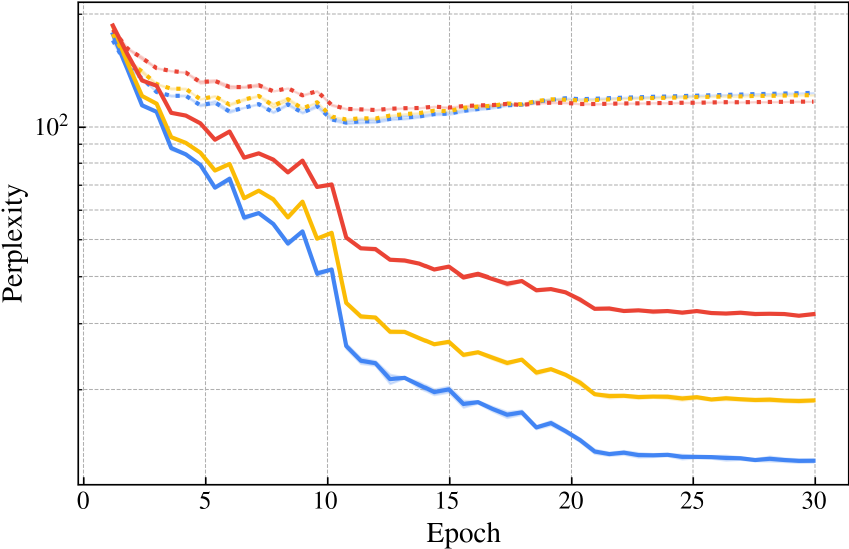}
        \caption{3 GRU Layers with 1024 Hidden Units}
        \label{fig:ptb3L1024}
    \end{subfigure}

    \caption{Comparison of Training and Validation Perplexity across Different Model Configurations on Penn Treebank. Each subfigure represents a different network configuration, varying in the number of layers and hidden size.}
    \label{fig:ptb_all}
\end{figure}

Figure~\ref{fig:ptb_all} presents the training and validation perplexities for various GRU configurations trained on the PTB dataset. Each subfigure illustrates a distinct network configuration varying in the number of layers (1 and 3) and the hidden dimension (256 and 1024 units), comparing BPTT, FT-BPTT and DSF.

From these results, we observe that BPTT consistently yields the lowest validation perplexity across all configurations, as expected, since it provides exact gradient computations. However, DSF performs remarkably close to BPTT, in both training and validation perplexities, 
    consistently outperforming FT-BPTT.

We also observe that generally, deeper (1 to 3 layers) and wider (256 to 1024 units) networks achieve lower perplexities across optimization methods. However, the benefits are nuanced: for the deepest and largest configuration (Figure~\ref{fig:ptb3L1024}), training instabilities appear to impact both BPTT and DSF, resulting in clear overfitting. Interestingly, in this scenario, the validation perplexities for DSF and BPTT deteriorate, crossing the FT-BPTT curve. This highlights the importance of carefully balancing model complexity and gradient propagation strategies.

\subsection{Impact of Model Size on Wikitext-103} 
\begin{figure}[ht]
    \centering
    \includegraphics[width=\linewidth]{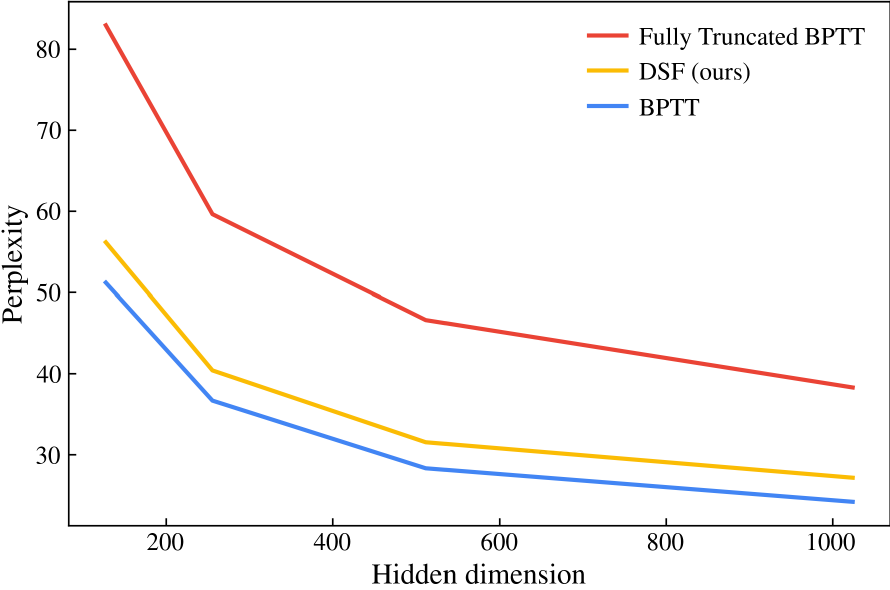}
    \caption{Comparison of Validation Perplexity across Different Hidden Dimension Sizes on Wikitext-103. The dimension size varies from 128 to 1024.}
    \label{fig:dim_experiments}
\end{figure}

In Figure~\ref{fig:dim_experiments}, we analyze the performance (measured by perplexity) across a variety of hidden dimensions for models trained on the Wikitext-103 dataset. Here, the hidden dimension $d$ varies significantly, and the results further underscore DSF's effectiveness.
First, while BPTT consistently achieves the lowest perplexity for each model size, DSF closely follows, significantly outperforming FT-BPTT across all tested model dimensions.
Furthermore, the gap between DSF and BPTT seems to stay constant with increasing hidden dimension, indicating that DSF maintains its effectiveness even in high-dimensional settings.

\subsection{Comparison Across Recurrent Architectures} 
\begin{figure}[ht]
    \centering
    \includegraphics[width=\linewidth]{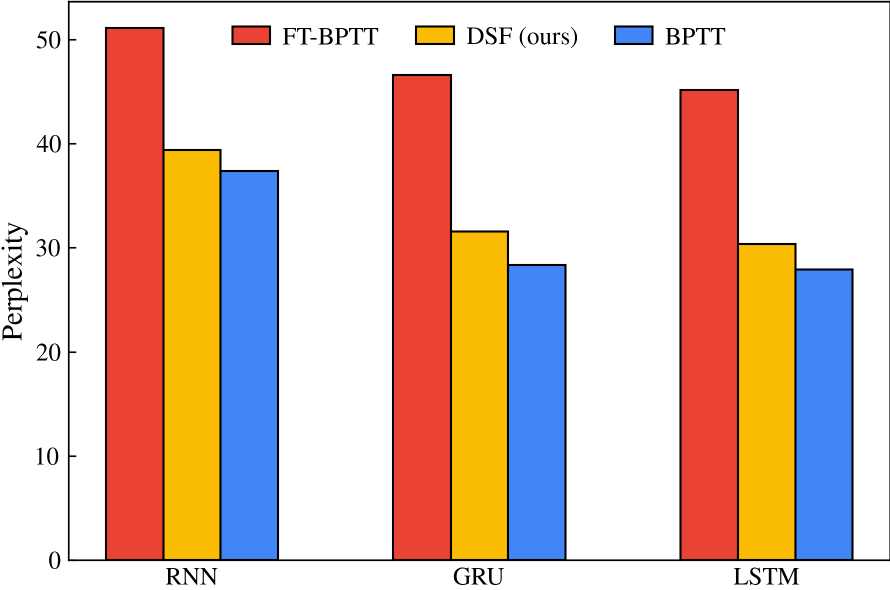}
    \caption{Comparison of Best Validation Perplexity across Different RNN Types on Wikitext-103. The models include standard RNNs, GRUs, and LSTMs of 
    3 layers with hidden size 512, trained with BPTT, DSF, and FT-BPTT.}
    \label{fig:rnn_types}
\end{figure}

Figure~\ref{fig:rnn_types} summarizes the best validation perplexity across different recurrent network architectures, including vanilla RNNs, GRUs, and LSTMs trained on Wikitext-103. 
We observe that as expected, gated RNNs consistently outperform simple RNN architectures, confirming their superior ability to capture long-range dependencies.
More interestingly, DSF consistently outperforms FT-BPTT, highlighting its capability to retain essential temporal gradient information despite significant simplifications.

DSF demonstrates effectiveness across different RNN types (vanilla RNN, GRU, and LSTM), maintaining performance close to BPTT. In the case of gated RNNs, a potential synergy between the gating mechanisms and the diagonal approximation of the feedback matrix could be at play, contributing to the competitive performance of DSF. 
However, in the case of vanilla RNNs, the gap between DSF and BPTT is also very small, which is counterintuitive except if considering that the memory mechanisms of RNNs are essentially diagonal.

\subsection{Effect of Network Depth (Layer Count)} 

Table~\ref{tab:optimization_methods} compares the best validation perplexities across various network depths (1, 3, 6, and 12 layers) on Wikitext-103.
\begin{table}[h]
    \setlength{\tabcolsep}{5pt}
    \centering
    \caption{Comparison of the best validation perplexity for models of different depths trained with BPTT, FT-BPTT, and DSF on Wikitext-103.}
    \label{tab:optimization_methods}
    \begin{tabular}{l|c|c|c|c}
        \toprule
        \textbf{Nb. of Layers} (Nb. Params) & \textbf{1} (20.3M) & \textbf{3} (28.2M) & \textbf{6} (40.0M) & \textbf{12} (63.7M) \\
        \midrule
        \textbf{BPTT}      & 35.59 & 28.32 & 25.44 & 23.73 \\
        \textbf{FT-BPTT}   & 56.61 & 46.60 & 41.35 & 36.59 \\
        \hline
        \textbf{DSF}       & 37.42 & 31.54 & 29.62 & 28.99 \\
        \bottomrule
    \end{tabular}
\end{table}
 We observe that as the depth increases, the perplexity consistently decreases for all optimization methods, confirming that deeper networks improve performance.
Furthermore, while BPTT achieves the best perplexity scores, DSF remains competitive across all tested depths. Notably, DSF seems to scale efficiently with depth, maintaining performance close to BPTT even in deep architectures.
FT-BPTT shows notably worse performance, highlighting that complete truncation of temporal gradient propagation substantially degrades model accuracy. DSF, which incorporates structured temporal feedback, substantially outperforms FT-BPTT and narrows the gap with BPTT significantly, even in deeper architectures.

\subsection{Effect of Sequence Length}
We report in Table~\ref{tab:seq_length} the validation perplexity of models trained on Wikitext-103 with sequence lengths ranging from 256 to 1024.

The results are overall consistent for all sequence lengths, with BPTT achieving the lowest perplexity, followed by DSF and FT-BPTT. DSF maintains competitive performance across different sequence lengths, demonstrating its robustness and scalability even in longer sequences.
\begin{table}[ht]
    \setlength{\tabcolsep}{8pt}
    \centering
    \caption{Validation perplexity comparison of models trained on Wikitext-103 with varying sequence lengths.}
    \label{tab:seq_length}
    \begin{tabular}{l|cccc}
        \toprule
        \textbf{Sequence Length} & \textbf{256} & \textbf{512} & \textbf{1024} & \textbf{2048} \\
        \midrule
        \textbf{BPTT} & 28.32 & 27.14 & 26.75 & 26.52 \\
        \textbf{FT-BPTT} & 46.60 & 45.95 & 46.75 & 46.52 \\
        \textbf{DSF} & 31.54 & 30.22 & 30.17 & 30.64 \\
        \bottomrule
    \end{tabular}
\end{table}

\subsection{Comparison with other Architectures}
We present a comparative evaluation of different sequence modeling architectures in Table~\ref{tab:perplexity_comparison}, measuring their validation perplexity on Wikitext-103 under a similar parameter budget. Note that all models have a transformer-like architecture, with only the attention layer replaced by either an RNN or an SSM.
The transformer model uses the GPT-2 architecture \cite{radford2019language} while the SSM model the EMA layer from MEGA \cite{ma2023mega}, which has a diagonal kernel.
\begin{table}[]
    \setlength{\tabcolsep}{8pt}
    \centering
    \caption{Validation perplexity comparison of Attention-based (Transformer) and State-Space Model (SSM) trained via BP, with RNN architectures trained via Backpropagation Through Time (BPTT) and Diagonal State Feedbacks (DSF) methods on Wikitext-103. Each architecture uses a comparable parameter budget, ensuring a fair comparison.}
    \label{tab:perplexity_comparison}
    \begin{tabular}{l|cccc}
            \toprule
           & \textbf{Attention} & \textbf{SSM}   & \multicolumn{2}{c}{\textbf{RNN}}  \\
           &           &       & BPTT        & DSF        \\
            \midrule
    \textbf{Perplexity}    & 24.58     & 33.19 & 28.32       & 31.54      \\
    \textbf{Params} & 25.9M     & 28.2M & 28.2M & 28.2M \\
            \bottomrule
    \end{tabular}
\end{table}

We find that as expected, the attention-based model attains the lowest perplexity of 24.58, outperforming all other architectures. This result is consistent with the dominance of Transformers in language modeling tasks, as self-attention enables capturing long-range dependencies efficiently. However, this comes at the cost of higher memory and computational requirements during inference, which are not reflected in the perplexity metric alone.

The RNN trained with full BPTT achieves a competitive perplexity of 28.32, demonstrating the advantage of exact gradient computations over alternative approximation methods. This remains the best-performing non-attention-based model, highlighting that precise temporal credit assignment still yields superior modeling capability.

The DSF-trained RNN attains a perplexity of 31.54, closely trailing behind BPTT despite using a fixed diagonal feedback matrix for gradient propagation. This indicates that while DSF approximates the full backpropagation gradients, it still captures sufficient temporal dependencies to yield strong performance, overperforming the SSM model which achieves a perplexity of 33.19. This suggests that, under comparable parameter budgets, SSMs may not always generalize as effectively as recurrent architectures in standard language modeling settings. The fact that DSF surpasses the SSM approach suggests that structured, fixed-feedback approximations can be more computationally efficient while retaining strong predictive power. This finding underscores the robustness of DSF's diagonal approximation, especially when compared to trained structured transition matrices used in SSMs.

The results confirm that while BPTT remains the most accurate training method, DSF provides an attractive alternative by striking a balance between computational efficiency and modeling performance. This makes DSF particularly suitable for scenarios where computational constraints are a concern, such as deployment on resource-limited devices.

\subsection{Summary of Observations}

The experimental evaluation highlights the following strengths of DSF:

\begin{itemize}
    
    \item \textbf{Structured feedback advantage}: DSF consistently surpasses Fully Truncated BPTT, highlighting that structured, even approximate, gradient propagation substantially improves training effectiveness.

    \item \textbf{Efficiency}: DSF achieves near-BPTT performance with significantly reduced computational complexity.
    
    \item \textbf{Scalability}: As model size and depth increase, DSF does not drift away from BPTT, demonstrating effectiveness in scaling to larger networks.

    \item \textbf{Competitive with State-Space Models (SSMs)}: Under equivalent parameter budgets, recurrent models trained with DSF can outperform SSMs, underscoring their competitiveness, even with approximated gradients.

\end{itemize}

\section{Discussion and Future Directions}\label{sec:discussion}

Our results demonstrate that the proposed Diagonal State Feedbacks (DSF) offers a compelling trade-off between computational efficiency and model performance. By constraining the feedback matrix to a fixed diagonal form, DSF significantly reduces the complexity and memory footprint associated with traditional Backpropagation Through Time (BPTT). 
Despite this simplification, empirical analyses show that DSF closely approaches the performance of BPTT across various network architectures, sequence lengths, and model sizes, consistently outperforming fully truncated BPTT.

One intriguing observation from our experiments is that the performance gap between DSF and standard BPTT seems to not depend on the depth or width. DSF thus appears effective for larger or deeper architectures, suggesting its usefulness to train large-scale models.
Several avenues merit further investigation to extend and refine DSF:
\begin{itemize}
    \item Adaptive feedback matrices: While our fixed diagonal matrix already provides significant computational and memory benefits, introducing controlled adaptability—such as updating the diagonal feedback entries occasionally or using meta-learning strategies—could further improve performance while maintaining computational advantages.
    
    \item Structured non-diagonal extensions: Exploring other structured approximations, such as low-rank plus diagonal or orthogonal matrices, could bridge the gap between diagonal approximations and dense matrices. Such structured alternatives might better capture complex temporal dependencies, potentially yielding superior modeling capacity at moderate additional cost.

    Investigating initialization strategies inspired by recent state-space model parameterizations (\textit{e.g.}, HiPPO, S4, S5) could further optimize the performance of DSF. Structured initialization might provide a beneficial inductive bias for modeling particular temporal dynamics or capturing long-term dependencies.
    However, underwhelming first experiments with such matrices indicate that finding suitable initializations might prove to be difficult.

    \item Theoretical analysis of gradient approximations: Further theoretical investigation into the impact of diagonal approximation on gradient propagation dynamics and stability could enhance understanding of when and why DSF performs effectively. These insights might guide practical design choices and inform new classes of simplified gradient methods for recurrent models.
\end{itemize}

Overall, DSF proposes a compromise between simplicity, scalability, and performance. Our empirical findings clearly indicate that with carefully designed structure and initialization, simplified feedback approximations can significantly reduce computational complexity without sacrificing predictive capability. These promising results pave the way for further exploration of structured, computationally efficient gradient propagation methods in recurrent neural networks. 

\section{Conclusion}\label{sec:conclusion}

In this paper, we introduced the Diagonal State Feedbacks (DSF) method, a novel approach to efficiently training recurrent neural networks by approximating the recurrent gradient computations using fixed diagonal feedback matrices. Our extensive empirical evaluations demonstrate that DSF significantly reduces both computational complexity and sequentiality compared to traditional Backpropagation Through Time (BPTT), while closely matching its predictive performance. The proposed method consistently outperforms fully truncated BPTT and remains competitive with more sophisticated models such as transformers and structured state-space models under comparable parameter budgets.

Our results highlight the practical effectiveness of DSF across different architectures, dataset scales, and model complexities, showcasing its versatility as a computationally efficient alternative to BPTT. This makes DSF particularly appealing for training large-scale recurrent models in resource-constrained or latency-sensitive environments.

Future research directions include exploring adaptive feedback structures, investigating alternative structured feedback matrices, and deepening theoretical understanding of these approximations. Ultimately, DSF offers a compelling balance between performance and efficiency, opening promising avenues for practical deployment of deep recurrent architectures.


\begin{credits}

\subsubsection{\discintname}
The authors have no competing interests to declare that are
relevant to the content of this article. 
\end{credits}

%
%
%
\bibliographystyle{splncs04}
\bibliography{biblio}
%
\newpage

\appendix

\section{Experimental Setup}

\subsection{Datasets}
\subsubsection{Penn Treebank}
The Penn Treebank dataset is a widely used benchmark for language modeling. It consists of 912k training tokens, and 131k validation tokens. The vocabulary size is 10k.

\subsubsection{Wikitext-103} The Wikitext-103 dataset is a larger version of the Wikitext-2 dataset, and is also a widely used benchmark for language modeling. It consists of over 100 million tokens from Wikipedia articles. We use a vocabulary size of 16k.

\subsection{Models}

\subsubsection{RNNs}
\begin{itemize}
  \item Vanilla RNN: The simplest form of RNN, where the hidden state is updated as:
   $$\htt = \sigma(W_1 \xt + W_2 \htt[t-1] + b)$$
  \item GRU \cite{cho2014learning}: A gated RNN, where the hidden state is updated as:
  \begin{align*}
    \mathbf{z}_t &= \sigma(W_z \xt + U_z \htt[t-1] + b_z) \\
    \mathbf{r}_t &= \sigma(W_r \xt + U_r \htt[t-1] + b_r) \\
    \htt &= (1 - \mathbf{z}_t) \odot \htt[t-1] + \mathbf{z}_t \odot \tanh(W_h \xt + U_h (\mathbf{r}_t \odot \htt[t-1]) + b_h)
  \end{align*}
  \item LSTM~\cite{hochreiter1997long}: A gated RNN with a short-term memory cell $\mathbf{c}_t$, and a long term memory cell $\mathbf{h}_t$. The hidden state is updated as:
  \begin{align*}
    \mathbf{i}_t &= \sigma(W_i \xt + U_i \htt[t-1] + b_i) \\
    \mathbf{f}_t &= \sigma(W_f \xt + U_f \htt[t-1] + b_f) \\
    \mathbf{o}_t &= \sigma(W_o \xt + U_o \htt[t-1] + b_o) \\
    \mathbf{c}_t &= \mathbf{f}_t \odot \mathbf{c}_{t-1} + \mathbf{i}_t \odot \tanh(W_c \xt + U_c \htt[t-1] + b_c) \\
    \htt &= \mathbf{o}_t \odot \tanh(\mathbf{c}_t)
  \end{align*}
  While an LSTM usually only outputs the short-term memory cell $\mathbf{c}_t$, we instead use the concatenation of both $\mathbf{c}_t$ and $\mathbf{h}_t$ to improve the performance of the model.
\end{itemize}

\subsubsection{Transformers}
We take inspiration from the GPT-2 model \cite{radford2019language} and use a transformer architecture with a stack of $L$ layers. Each layer consists of a multi-head self-attention mechanism followed by a feed-forward neural network. Both the attention mechanism and the feed-forward network are wrapped by a residual connection and their inputs are normalized using layer normalization. The feed-forward network's intermediate dimension is 4 times the hidden dimension of the model.

\subsubsection{SSMs} We use the SSM as used in the MEGA model \cite{ma2023mega}. It is a diagonal SSM akin to S4D but with a different initialization scheme, and real-valued. 

\subsection{Training}
\subsubsection{Setup}
All our experiments are run on single A100 GPUs. We use the Adam optimizer with the default values for momentum and second order momentum. We use TF32 precision to speed up training with no performance degradation.

\subsubsection{Hyperparameters}
We report the hyperparameters used for training the models in Table~\ref{tab:hyperparameters}.
\begin{table}[h]
  \centering
  \begin{tabular}{l|c|c}
    \toprule
    \textbf{Dataset} & \textbf{Penn Treebank} & \textbf{Wikitext-103} \\
    \midrule
    \textbf{Layers} & 3 & 3 \\
    \textbf{Hidden dimension} & 256 & 512 \\
    \textbf{Vocabulary size} & 10k & 16k \\
    \textbf{Skip connections} & \cmark & \cmark \\
    \textbf{Transformer-like} & \xmark & \cmark \\
    \textbf{Context length} & 64 & 256 \\

    \hline
    \textbf{Epochs} & 30 & 5 \\
    \textbf{Batch size} & 128 & 128 \\
    \textbf{Learning rate} & 1e-3 & 1e-3 \\
    \textbf{Weight decay} & 1e-4 & 0 \\
    \textbf{Scheduler} & Step ($\times 0.1$ at 10, 20) & Cosine \\

    \bottomrule
    \end{tabular}
  \caption{Hyperparameters for the models used in the experiments, unless stated otherwise.}
  \label{tab:hyperparameters}
\end{table}

\end{document}